\documentclass{article}

\usepackage[utf8]{inputenc}
\usepackage{graphicx}
\usepackage{algorithm}
\usepackage{algpseudocode}
\usepackage{amssymb}
\usepackage{amsmath}
\usepackage{subfig}
\usepackage{xcolor}
\usepackage{eurosym}
\usepackage{nameref}

\usepackage[style=numeric,backend=biber,giveninits=true,uniquename=init,sorting=none,natbib=true,sortcites=no]{biblatex}
\DeclareMathOperator{\Betad}{Beta}

\title{Inferring perceptual decision making parameters from behavior in production
and reproduction tasks}
\author{Nils Neupärtl$^{\text{a},\text{b}}$ \qquad Constantin A. Rothkopf$^{\text{a},\text{b}}$}
\date{$^{\text{a}}$Centre for Cognitive Science, TU Darmstadt, Darmstadt, Germany \\ \vspace{0.1cm}
$^{\text{b}}$Institute of Psychology, TU Darmstadt, Darmstadt, Germany}

\begin{document}

\maketitle

\begin{abstract} 
	Bayesian models of behavior have provided computational level explanations in a range of psychophysical tasks. One fundamental experimental paradigm is the production or reproduction task, in which subjects are instructed to generate an action that either reproduces a previously sensed stimulus magnitude or achieves a target response. This type of task therefore distinguishes itself from other psychophysical tasks in that the responses are on a continuum and effort plays an important role with increasing response magnitude. Based on Bayesian decision theory we present an inference method to recover perceptual uncertainty, response variability, and the cost function underlying human responses. Crucially, the cost function is parameterized such that effort is explicitly included. We present a hybrid inference method employing MCMC sampling utilizing appropriate proposal distributions and an inner loop utilizing amortized inference with a neural network that approximates the mode of the optimal response distribution. We show how this model can be utilized to avoid unidentifiability of experimental designs and that parameters can be recovered through validation on synthetic and application to experimental data. Our approach will enable behavioral scientists to perform Bayesian inference of decision making parameters in production and reproduction tasks.
\end{abstract}

\section{Introduction}
Psychophysics has developed a variety of experimental paradigms to measure human decision making under perceptual uncertainty. 
Different experimental paradigms can be distinguished by the specifics of how stimuli are generated, presented, and which responses are required by participants. 
One of the most common paradigms is the two-alternative forced choice task (2AFC) \citep{gescheider2013psychophysics,stevens1958problems,wichmann2018methods,stuttgen2011mapping}, which confronts the subject with a binary decision regarding the property of a stimulus and thereby allows the experimenter to draw conclusions about the subject's perceptual sensitivity given the stimulus and the task conditions. 
Other experimental paradigms include the two-interval forced choice task (2IFC), where stimuli are presented sequentially, and the yes-no task, in which the subject needs to detect a stimulus, which is either present or absent \citep{gescheider2013psychophysics,stevens1958problems,wichmann2018methods,stuttgen2011mapping}.\\

Because all these tasks involve binary responses,
they differ fundamentally from other psychophysical paradigms such as  
production and reproduction task. While production tasks ask subjects to generate a graded response with a target magnitude, in reproduction tasks subjects first sense a stimulus magnitude and are then instructed to reproduce the sensed stimulus magnitude.
The distinguishing factor here is that actions can be taken on a continuous scale and thus, in addition to the purely perceptual uncertainty, action variability is introduced into the overall noise manifested in decisions.
Classic examples of a production task include walking to targets using visual cues \parencite{harris2000visual,mittelstaedt2001idiothetic} and time estimation tasks after stimuli onset \parencite{wild2009feedback,miltner1997event}
and of a reproduction task are path length reproduction \parencite{berthoz1995spatial,mcnaughton2006path,petzschner2011iterative}, time interval reproduction \parencite{fraisse1984perception,buhusi2005makes} and force reproduction \parencite{shergill2003two,walsh2011overestimation,onneweer2015force}.\\

Bayesian models combining prior beliefs with sensory measurements have been successfully applied to provide computational level explanations of human behavior in perceptual decision tasks \citep{knill1996perception}. Very often, these studies employ a particular psychophysical paradigm, to measure perceptual uncertainties. Exemplary studies on human cue integration such as \citet{ernst2002humans} and \citet{knill2003humans} first measured the uncertainties of individual cues with classic psychophysical tasks. In both cases, the models assumed Gaussian distributions describing perceptual noise or more generally, perceptual uncertainties. Similarly, experimental paradigms involving continuous actions in visuomotor behavior have often also used Gaussian distributions to capture response variability. Exemplary studies such as K\"ording and Wolpert \parencite{kording2004bayesian} and \citet{trommershauser2003statistical} represented spatial distributions of manual response variability with normal distributions.\\

According to Bayesian decision theory, the optimal decision depends not only on the combination
of prior beliefs about unobserved environmental parameters with the likelihood of these parameters given sensory measurements 
as represented by the
inferred posterior, but additionally takes the response variability introduced by the necessary action into account. The loss function encompasses the costs and benefits for deviating from the true target value, which is the consequence of response variability. 
Broad experimental evidence suggests, that the nervous system takes this variability into account when carrying out movements \parencite{kording2007decision,harris1998signal,o2009dissociating}.
If the involved uncertainties are normally distributed and the cost function is symmetric, then selecting the optimal response leads to an action that corresponds to the mean of the posterior distribution, which coincides with the mode and the median in case of normal distributions. But in case of asymmetric cost functions and skewed probability distributions describing perceptual uncertainties and action variability, these quantities may interact in non-trivial ways resulting in optimal responses, which can be systematically biased. Importantly, while human cost functions in psychophysical tasks have been measured, they are often conveniently assumed to be symmetric \citep{kording2004loss}.\\

Here we introduce a computational approach to infer the parameters in Bayesian decision models of production and reproduction tasks in the spirit of rational analysis \parencite{simon1955behavioral,griffiths2015rational}. 
The model employs log-normal distributions both for the perceptual uncertainty, as this accommodates well known perceptual Weber-Fechner phenomena \parencite{battaglia2011haptic,dehaene2003neural}
, and for the response variability observed in humans 
\parencite{harris1998signal,van2004role,hamilton2004scaling}.\\

Specifically, we infer individual’s perceptual uncertainty, response variability, and their subjective cost function based on the responses in production and reproduction tasks. 
Importantly, these types of tasks require a graded response with different magnitudes. Thus, behavior balances the trade-off between task fulfillment and effort, leading to possibly non-trivial response distributions of undershoots and overshoots, which are ubiquitous in human psychophysical experiments.
We show that this framework is able not only to recover the parameters in Bayesian decision models for diverse cost functions but additionally can be utilized to guide designing psychophysical experiments.

\section{Methods}
\begin{figure}
	\centering
	\includegraphics[width=\textwidth]{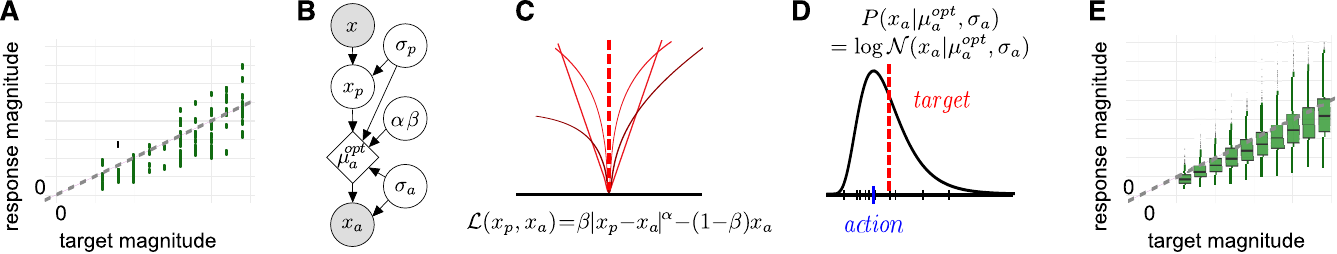}
	\caption{\textbf{Inference model}. \textbf{A} Typical behavioral data in a production or reproduction task. \textbf{B} Bayesian network of response generation from the perspective of the researcher. \textbf{C} Examples of parameterized costfunctions including effort. \textbf{D} Log-normal response distribution. \textbf{E} Simulated responses using inferred model parameters.}
	\label{fig:data}
\end{figure}

\subsection{Continuous Cost Optimized Parameter Inference for decision making tasks}
Here we present a method to infer parameters from subject's behavior in decision making tasks involving a continuous response, which includes production and reproduction tasks. We employ Markov Chain Monte Carlo (MCMC) and optimization based on a neural network approximation to infer posterior distributions describing perceptual uncertainty, action variability and parameterized cost functions, including costs for action outcomes and response efforts. These parameters can be inferred on an individual-by-individual basis for for a wide range of continuous response tasks and data. By using log-normal distributions for perceptual uncertianty and action variability we constrain values to a strictly positive range and importantly reproduce Weber-Fechner like phenomena, i.e. variability that increases with stimulus magnitude.

\subsection{Bayesian observer model}
We first assume that the experimenter's stimulus is a continuous random variable $x \in \mathbb{R}^+$. Perception leads to an internal variable $x_p$, which we model as having a log-normal distribution with variance $\sigma_p$ resulting in the distribution $P(x_p|\mu_p, \sigma_p)~$:
\begin{eqnarray}
\displaystyle{P(x_p|\mu_p, \sigma_p) = \log\,\mathcal{N}(x_p | \mu_p, \sigma_p) \;\;\; \textrm{with} \;\; \mu_p = \log(x)+\sigma_p^2}
\label{eq:perception_log-normal}
\end{eqnarray}
where $\log\,\mathcal{N}(x_p | \mu_p, \sigma_p)$ is a log-normal distribution with its mode at the actual target $x$. This modeling choice accommodates the known scaling of uncertainty in the human sensory system, commonly referred to as Weber-Fechner scaling \parencite{battaglia2011haptic,neupaertl2020intuitive}. The perceptual uncertainty can either be measured independently through experimental means \parencite{wichmann2018methods} or it can be inferred given behavioral data. In the former case its value is fixed and in the latter case a hyperprior incorporating prior knowledge about the possible range of its magnitude can be used with $\sigma_{p}\sim \log\mathcal{N}(\mu, \sigma^2)$.

In order to act optimally given a task and the perceived magnitude of the stimulus, human subjects are assumed to try to minimize their costs as described by Bayesian decision theory (BDT). 
In production and reproduction tasks, subjects aim at getting as close as possible to a target magnitude with their response.
Therefore, in these tasks costs depend on how close the action leads to the actual target. 
It is commonly assumed that subjects evaluate outcomes based on a function of their absolute error \parencite{kording2004loss,wolpert2012motor}, i.e. that they assess the value of an action on a continuous scale based on the absolute distance from the target, regardless of whether they fall above or below the desired value. Such cost functions are symmetrical around the target value and can be formalized as:
\begin{eqnarray}
\displaystyle{\mathcal{L}^{'}(x_p,x_a) = |x_p-x_a|^{\alpha}}
\label{eq:cost_simple}
\end{eqnarray}
with the target value $x$, the value of a single action outcome $x_a$ and an exponent $\alpha$ describing the shape of the cost function, e.g. with values of $\alpha$ at 1 or 2 cost functions correspond to the $L_1$ or hinge-loss and $L_2$ or quadratic cost functions, respectively.

However, 
production and reproduction experiments require subjects to produce a response that increases in magnitude with the sensed magnitude of the stimulus. 
Thus, the task-dependent description of costs does not yet suffice, since actions themselves are already associated with effort and thus costs. These costs can now be included in the cost function term relative to the costs for the outcome of the action:
\begin{eqnarray}
\displaystyle{\mathcal{L}(x_p,x_a) = \beta \, |x_p-x_a|^{\alpha} + (1-\beta) \, x_a}
\label{eq:cost_effort}
\end{eqnarray}
with $\beta$ governing the trade-off between task costs and production effort. That is, if $\beta$ is close to $1$, actions are dominated by the costs due to the expected errors whereas if $\beta$ is close to $0$, actions are instead dominated by the effort cost due to the response action.
Accordingly the factor $1-\beta$ describes the raye of effort increase for responses at higher magnitudes. A suitable prior over the parameter $\beta$ can be obtained using the beta distribution $\beta \sim \Betad(a_{1},a_{0})$.
The parameter $\alpha$ again determines the shape of the task dependent costs and its hyperprior is chosen as a log-normal distribution with $\alpha \sim \log\mathcal{N}(\mu, \sigma^2)$.

The actions themselves are in turn affected by variability, which is also known to be signal dependent and the nervous system takes this variability into account when carrying out movements. 
Therefore, the optimal action takes into account the perceptual uncertainty, the expected action variability and applies the costs stemming from the task's goals and the effort in producing the response. This can be expressed as the solution to the minimization of the overall expected loss:
\begin{eqnarray}
\displaystyle{\mu_a^{opt} = \min_{\mu_a\in \mathcal{R}^*_+} \int_{0}^{\infty}\int_{0}^{\infty}  \mathcal{L}(x_p,x_a|\alpha,\beta) \, p(x_a|\mu_a,\sigma_a) \, p(x_p|\mu_p,\sigma_p) \,dx_p \,dx_a }.
\label{eq:cost_minimization}
\end{eqnarray}
Algorithmically, we use amortized inference to approximate response distribution using a regression neural network trained with numerically solved samples.
The action variability is already included in the optimization problem to obtain the optimal action while considering and compensating for motor noise as well. This leads to log-normal response distribution with a mode that reflects individual task and action specific costs while including perceptual and action associated variability:
\begin{eqnarray}
\displaystyle{P(x_a|\mu_a^{opt}, \sigma_a) = \log\mathcal{N}(x_a | \mu_a^{opt}, \sigma_a)}
\label{eq:action_log-normal}
\end{eqnarray}

\begin{figure}
	\centering
	\includegraphics[width=1\textwidth]{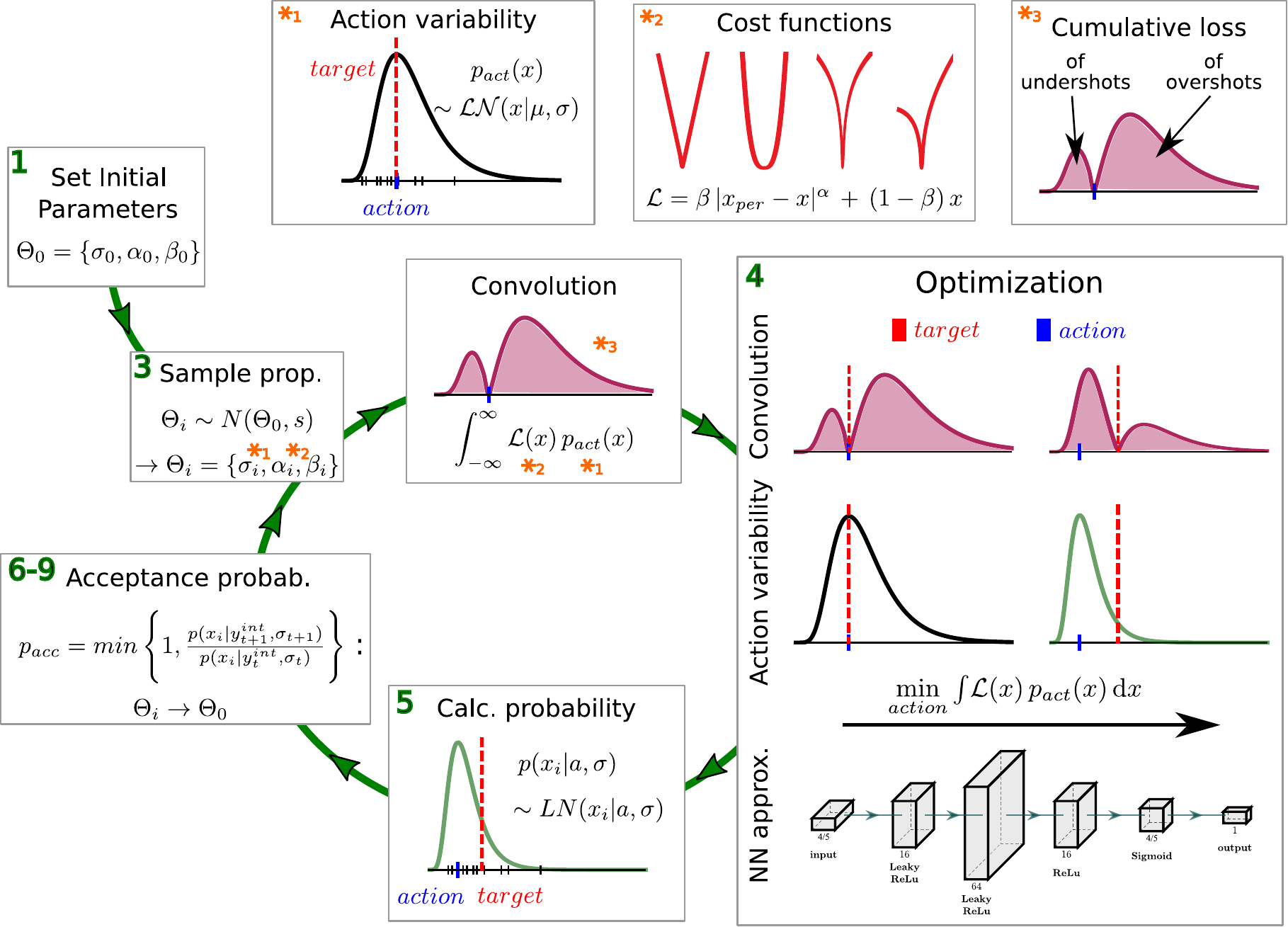}
	\caption{\textbf{Visualization of cost inference algorithm.} Numbers reflect the steps in Algorithm \ref{pseudo:UC-MCMC} while red arrows indicate the process direction. The chain starts at 1) with the initial parameter for action variability and cost function. In 3) based on these parameters and their associated stepsizes a first new proposal is drawn from a Gaussian distribution. These values lead to a response distribution, a cost function and the result of their product the weighted distribution. This distribution has its mode initially at the target position - that e.g. is the value to be reproduced in a reproduction task. However, depending on the value of alpha and beta, this choice may not necessarily have the lowest cumulative cost and therefore is optimized in 4) with respect to the position of the mode of the response distribution $y_{int}$. 
	The optimization process of finding the best action given the constraints is approximated with a neural network shown in the bottom row for both known and unknown perceptual uncertainty.
	Input values are action variability $\sigma_a$, cost function coefficients $\alpha$ and $\beta$, the actual target $x$ and - if unknown - the perceptual uncertainty $\sigma_p$. Networks had six layers: an input layer with 4 or 5 units, layer one with 16 units and leaky ReLu activation (alpha at .1), layer two with 64 units and leaky ReLu activation (alpha at .05), layer three with 16 units and ReLu activation, layer four with 4 or 5 units and sigmoid activation, and layer 5 as a single unit output layer.
	In step 5) the likelihood of the data given $\sigma$ and $y_{int}$ is calculated as well as the prior probability of $\sigma$, $\alpha$ and $\beta$ for the initial parameters. 
	MH acceptance rule in steps 6-9).
	}\label{fig:concept}
\end{figure}

Due to the non-symmetrical form of $p(x_p)$ and $p(x_a)$ together with non-trivial cost functions, optimal responses do not necessarily have to aim for the actual target.  
Thus, the interplay of perceptual uncertainties, response variability, and complex asymmetric cost functions can explain biases and systematic deviations in subjects' responses.
Inference of these parameters given observed behavior can therefore quantitatively ascribe seemingly suboptimal behavior to objective task parameters and subjective costs and benefits. 
\begin{algorithm}[ht]
	\caption{Inference algorithm for production and reproduction tasks}
	\begin{algorithmic}[1]
		\State Draw chain startvalues $\Theta_0=[\alpha_0,\beta_0,\sigma^{p}_0,\sigma^{a}_0]$ from priors and MH-stepsize $s$ \vskip 2pt
		\For{iterations $i$ from $1$ to $N$} \vskip 2pt
		\State Draw new proposal $\Theta_{p}\sim N(\Theta_{i-1},s)$\vskip 2pt
		\State Approximate optimal action distribution parameter $\mu_a^{opt}$ via neural network\newline
		\indent $\Longrightarrow$ find $\mu$ given $\alpha$, $\beta$, $\sigma_{p}$, $\sigma_{a}$ according to eq.\ref{eq:cost_minimization} \vskip 2pt
		\State Calculate posterior probability of actual responses $P(x_a^{act}| \mu_a^{opt}, \sigma_a)$ \vskip 2pt
		\If{ $\frac{p(proposal)}{p(previous)}$ > $r \sim U[0,1]$} \vskip 2pt
		\State Accept proposal $\Theta_i \Leftarrow \Theta_p$
		\Else
		\State Reject proposal $\Theta_i \Leftarrow \Theta_{i-1}$
		\EndIf
		\EndFor \vskip 2pt
	\end{algorithmic}
	\label{pseudo:UC-MCMC}
\end{algorithm}
The whole process is 
The pseudo algorithm for the complete inference process is shown
The complete inference process is summarized as pseudo algorithm \ref{pseudo:UC-MCMC} and additionally visualized in figure \ref{fig:concept}.\\

\subsection{MCMC with neural network approximated optimization}

To infer the aforementioned parameters we utilize Markov Chain Monte Carlo (MCMC) sampling together with a neural network, which approximates the parameters of the optimal response distribution given the sampled parameter values. Similar techniques for likelihood or posterior approximations via neural networks are also increasingly popular across various fields from reinforcement learning \parencite{wulfmeier2015maximum,hamrick2020role} and cognitive neuroscience \parencite{fengler2021likelihood} to population genetics \parencite{beaumont2002approximate}.
We start the Metropolis-Hastings algorithm by drawing samples from symmetrical proposal distributions for each parameter.
Given the resulting cost function, magnitude of the variability and the target one can calculate the optimal parameter of the response distribution $\mu_a^{opt}$, i.e. the optimal aiming point for the response. Given the asymmetry of a log-normal distribution, its mode does not have to coincide with the target. Accordingly, a subject could intentionally aim for values deviating from the target in an attempt to minimize her own costs. For example, 
in a task requiring responses involving  large effort, it may be more suitable to undershoot the target in order to obtain a good task-effort trade-off. 

\subsection{Network structure and training}
The optimization needs to be done for each data point in each MCMC iteration, which is costly for large data sets originating from experiments with continuous response variables. Therefore, we trained a neural network approximating this optimization, allowing to infer behavioral parameters on large data sets more efficiently. With the neural network yielding the ideal shift of the log-normal distribution's mode for each data point and the previously drawn action variability $\sigma_a$ one can calculate the likelihood of the data given the model and its parameters.\\

For inference of the optimization, we utilized a regression neural network (see figure \ref{fig:concept} step 4 \textit{'Optimization'}) with four hidden layers approximating the optimal position of the response distribution's mode given the perceptual uncertainty $\sigma_p$, the action variability $\sigma_a$, both cost function parameters $\alpha$, $\beta$ and the actual target $x$. 
The six layers are as follows: an input layer with, depending whether $\sigma_p$ is known or not, 4 or 5 units, layer one with 16 units and leaky ReLu activation (alpha at .1), layer two with 64 units and leaky ReLu activation (alpha at .05), layer three with 16 units and ReLu activation, layer four with 4 or 5 units and sigmoid activation, and layer five as a single unit output layer.\\

We trained several networks depending on whether the perceptual uncertainty was assumed to be known and thus could be fixed to a specific value or whether it was assumed to be unknown and therefore needed to be inferred, too. Training used 200 epochs for each neural network with fixed $\sigma_p$, only differing in the magnitude of perceptual uncertainty ($\sigma_p$ at .05 and .2 for the puck and beanbag task, see figure \ref{fig:puck_data_inference} \& \ref{fig:beanbag_data_inference}) and 600 epochs for the neural network with variable $\sigma_p$ (see figure \ref{fig:costInf_sigma_tile}). 
In both cases we used the mean-squared-error loss for training, early stop callbacks based on validation loss, and a validation split of 0.2. The training data consisted of 18,910 samples for networks with known and 123,752 samples for the network with unknown $\sigma_p$.
To obtain training data,
we sampled parameter values for $\alpha$, $\beta$, $\sigma_p$, $\sigma_a$ and target positions (4-par network: $\sigma_a$ .01 to 1, $\alpha$ .01 to 4, $\beta$ .5 to .99 and target .2 to 5; 5-par network: $\sigma_a$ .01 to 1, $\sigma_p$ .005 to 1, $\alpha$ .01 to 8, $\beta$ .5 to .99 and target .2 to 5).
For each resulting parameter set, we 
calculated the mode of the optimal action distribution, i.e. the distribution yielding the lowest costs, by solving equation \ref{eq:cost_minimization} numerically with Brent's method. 
Training was stable with a mean absolute error of $0.05$.
It should be pointed out, that using this network is predominantly a tool to speed up the optimization shown in equation \ref{eq:cost_minimization}.
Evaluation of each iteration's proposals is again performed via rejection sampling as in the traditional Metropolis-Hastings algorithm.\\

\section{Experiments}
In the experimental evaluations we first show how the proposed method can be used to investigate how behavioral responses such as undershoots and overshoots in potential experiments depend on both the experimental parameters and parameters describing a subject's uncertainty and cost function.
This can be utilized to adjust experimental parameters to facilitate inference of the model's parameters given observed behavior. 
We proceed to show how
the proposed method can be utilized to obtain posterior distributions over parameters describing behavior in production and reproduction tasks. 
Specifically, we evaluate the inference algorithm on synthetic data showing that
it is possible to recover individual posterior probabilities for the parameters governing perceptual uncertainties and parameter of the cost function based on empirical data. 
Finally, we apply our method to experimental data from two production tasks \parencite{neupaertl2020intuitive,willey2018long}, showing that 
the cost functions and action variability parameters can be recovered.

\subsection{Investigating the feasibility of task designs}
\begin{figure}[!t]
	\centering
	\includegraphics[width=\textwidth]{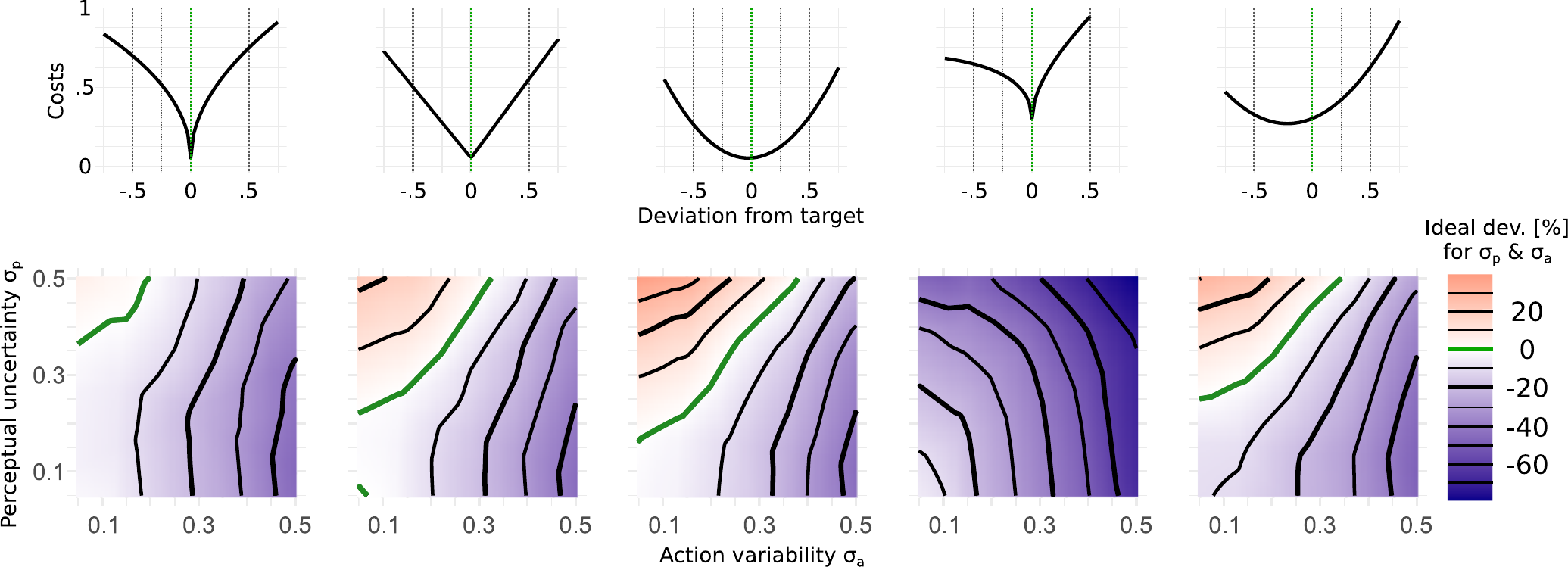}
	\caption{Tile plot showing the influence of perceptual uncertainty $\sigma_p$ and action variability $\sigma_a$ on the intentional target, i.e. mode of the log-normal() response distribution, for five different cost functions. Both axes range from .05 to .5 to display realistic values. 
	Color code describes whether actions are expected to match the target (green line), undershoot it (blue regions) or overshoot it (red regions).
	Deviations are presented on a percentage scale of the actual target.
	}
	\label{fig:costInf_sigma_tile}
\end{figure}

To investigate the expected behavior in an experiment, we simulate the responses arising from the Bayesian decision model for parametrically changing perceptual uncertainty and response variability. By assuming different cost functions including those that explicitly implement subjective costs for effort, we can investigate the resulting parameter ranges with associated undershoots and overshoots.
The corresponding parameter spaces can be visualized, as in figure \ref{fig:costInf_sigma_tile}.
These plots show the optimal aiming point for each combination of 
action variability (x-axis) and perceptual uncertainty (y-axis).
The optimal aiming point corresponds to 
the mode of the resulting response distribution and 
accordingly undershoots are colored in blue while overshoots are colored in red.
The combination of parameters resulting in an optimal aiming point coinciding with the perceived position of the target, i.e. an unbiased aiming point from the perspective of the experimenter, is marked in green.
Each column shows the relative position of optimal aiming points for different types of 
exemplary cost function: from symmetric cost functions (column 1-3) to asymmetric ones involving higher effort costs (column 4,5).\\

This allows basic conclusions about the influences of different magnitudes of the individual parameters: i) increasing action variability result in an increasing tendency to undershoot the target (see x-axis in figure \ref{fig:costInf_sigma_tile}), the consequence of avoiding costly overshoots arising from the heavy-tailed log-normal response distribution, ii) high values of perceptual uncertainty, on the other hand, can lead to overshoots (see y-axis), especially when action variability is low and $\alpha$ coefficients are high, see first three columns in figure \ref{fig:costInf_sigma_tile}. 
Consider for example the case of high perceptual uncertainty about the actual target and no action variability. Then, the responses falls exactly on the mean of the perceptual distribution for the squared loss. Thus, the experimenter would observe overshoots, since the actual target is the mode of this perceptual distribution. This effect is further enhanced by stronger subjective penalties on the response errors, i.e. larger values of alpha, see figure \ref{fig:costInf_sigma_tile} columns 1-3.\\

These theoretical implications give experimenters the opportunity to consider in advance effects of certain parameter adjustments. When designing an experiment, experimenters always set up a cost function, even if it is only implicit in the task description. By changing the task demands or by directly including an explicit cost function they can influence the potential parameter space for participants. Similarly, it can be useful to measure the perceptual uncertainty of subjects in advance or to manipulate it in certain task conditions in order to guide behavior in predictable ways.

\subsection{Inferring behavioral parameters from synthetic data}
\begin{figure}[H]
	\centering
	\includegraphics[width=\textwidth]{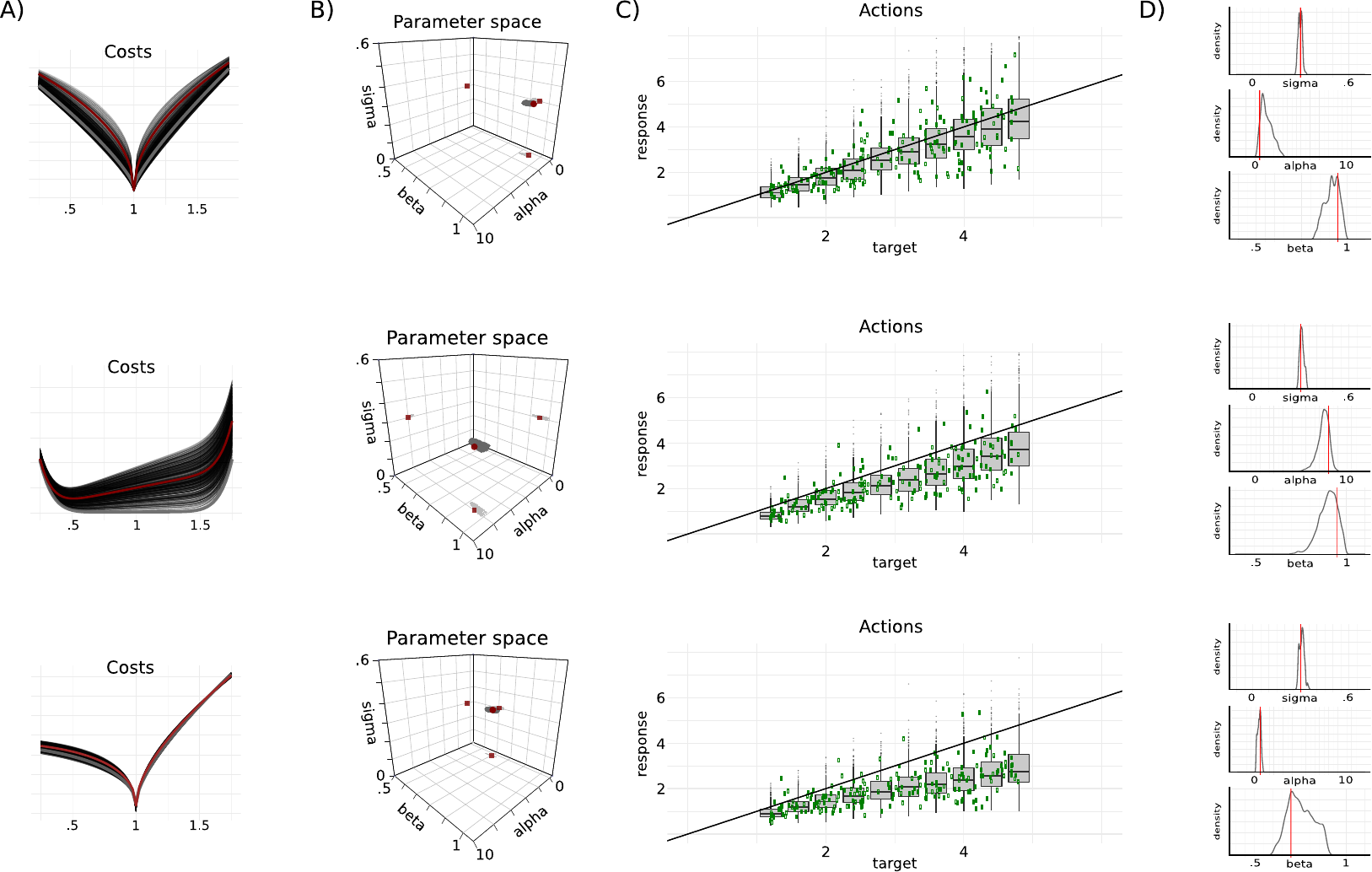}
	\caption{Response pattern of a generic reproduction task and its recovery for three different cost parameter settings. 
	(A) Recovered cost functions and ground truth cost function (dark red).
	(B) Three-dimensional space with the most likely recovered parameters in dark grey and ground truth values in dark red.
	(C) Generated responses based on ground truth parameters as green dots and predictions as box plots based on the most likely sample.
	(D) Posterior distributions for $\alpha$, $\beta$ and $\sigma_a$ parameters and ground truth marked by red vertical lines.
	}
	\label{fig:gen_data}
\end{figure}

We proceed to validate the proposed inference algorithm by showing numerical evaluations on synthetic data sets allowing to compare recovered parameters with ground truth.
In all simulations we used the priors $\beta \sim \Betad(10,2)$ and $\alpha \sim log\,\mathcal{N}(3.5, 2)$, initialized eight chains to explore the parameter space with 5,000 samples each, used the results to start a chain with 20,000 samples and optimized initial parameters and stepsize.
First, we demonstrate the ability of the algorithm to recover parameters from synthetic data and use the framework to generate predictions based on the best parameter setting. To do so we generated 200 data points, uniformly distributed on a continuous range (from 1.2 to 4.8, see generic responses as green data points in figure \ref{fig:gen_data} (C)). This allows a test under realistic conditions, as they occur e.g. in psychophysics tasks as time reproduction \parencite{birkenbusch2015octuplicate}, throwing \parencite{willey2018limited,willey2018long}, lifting or walking tasks \parencite{petzschner2011iterative,sun2004multisensory}. These values can now be thought of as target values that a subject tries to achieve with her actions and can be interpreted on trial by trial and subject by subject level. Here, we chose three different sets of data, which vary in their cost function parameters but neither in their range nor their uncertainties (all with $\sigma_p = .05$ and $\sigma_a = .3$). Thus, besides investigating the ability to recover values, influences of different cost functions on behavioral response patterns become more conspicuous.\\

Each row in figure \ref{fig:gen_data} shows one 
simulation of a particular subject with a particular cost function, the recovered parameters and resulting predictions. With clear prioritization of the task, i.e. a high task motivation (here, $\beta = .95$), and a concave functional dependence of the cost on the absolute distance ($\alpha = .5$), we can observe a cost function, almost symmetric around the target value, in the first row. Together with an intermediate action variability ($\sigma_a = .3$) and a low perceptual uncertainty ($\sigma_p = .05$), this nearly symmetric cost function leads to the largely unbiased behavior in (C) as green data points, which differ from the ideal line only through the increased action variability. Given the data it is now possible to reverse engineer the underlying parameters: (D) depicts the complete posterior distributions for $\sigma_a$, $\alpha$ and $\beta$, of which the five percent data points with the highest likelihood are additionally represented in the three-dimensional parameter space (B). In (A), the ground truth cost function (in darkred) and the cost functions belonging to the inferred parameters are plotted. The cost function belonging to the recovered parameter setting with the highest likelihood is used to generate synthetic responses shown as grey boxplot in (C) (10,000 data points distributed over ten targets). The same procedure is followed in the second and third row with modified cost functions: second row shows a flatter cost increase around the target ($\alpha = 8$) and thus a slightly increased influence of the action cost, despite unchanged $\beta$, resulting in the expected slight deviation of the behavior from the ideal line in data and prediction. In the third row however, $\alpha$ is again at $0.5$ as in the first row, only this time with a decreased value of $0.7$ for $\beta$ and thus creating higher costs for longer or greater actions. Deviations of the inferences were extremely small for the action variability $\sigma_a$ with modes at $0.3053$, $0.3034$ and $0.3091$ and RMS errors of $0.0124$, $0.0149$ and $0.0164$ for the three data sets, respectively, given the actual value of $0.3$. Inferred values for the $\alpha$ parameter show higher variability with modes at $0.8209$, $7.5105$ and $0.4910$ given their actual values at $0.5$, $8$ and $0.5$ and with RMS errors of $0.9928$, $0.8846$ and $0.2136$. Inferences for $\beta$ yield modes at $0.9161$, $0.9101$ and $0.7047$ with actual values at $0.95$, $0.95$ and $0.7$ and RMS errors of $0.0510$, $0.0706$ and $0.0885$. This validates the overall high precision in recovering the parameters under variability for a set of just 200 sampled data points.
\\

\begin{figure}[!ht]
	\centering
	\includegraphics[width=\textwidth]{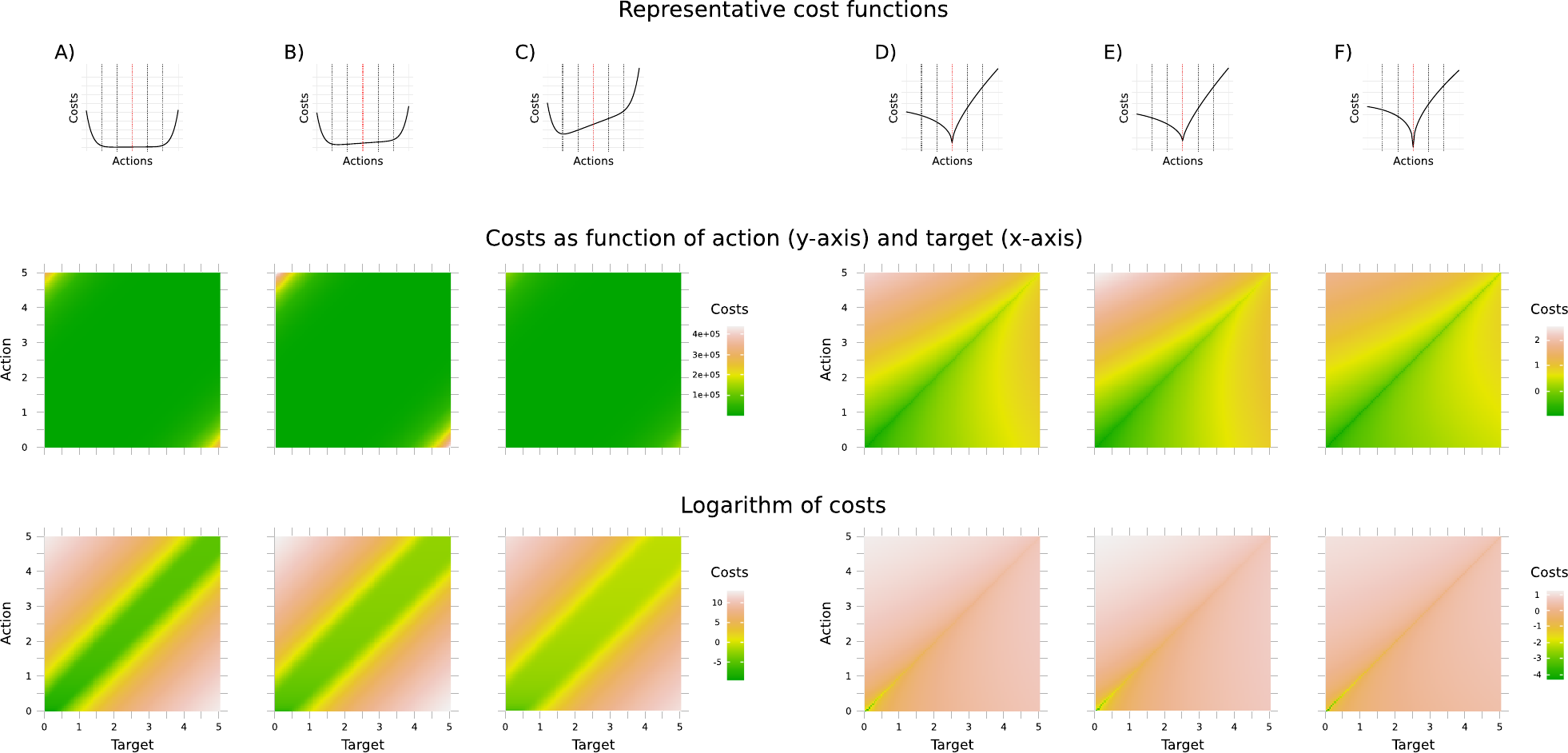}
	\caption{Comparison of three sampled cost functions with high posterior probability for each of the two simulated data sets in the second and third row in figure \ref{fig:gen_data}. Bottom row depicts the log value of costs to better visualize the steep gradients, especially for the left example. Left panel A-C): three exemplary cost functions with associated costs for the second case in figure \ref{fig:gen_data}.
	Right panel D-F): three exemplary cost functions with associated costs for the third case in figure \ref{fig:gen_data}.}
	\label{fig:costInf_2d_cost-tiles}
\end{figure}

For a more profound explanation and visualization of the resulting samples of cost function parameters we can display, besides the one-dimensional cost functions 
for a specified target (e.g. at 1 as in figure \ref{fig:gen_data}), 
costs as a function of both, potential targets and actions.
Figure \ref{fig:costInf_2d_cost-tiles} shows targets and actions in a range from $0$ to $5$ with corresponding costs as colored tile plots for samples of the two generic data sets in the second and third row of figure \ref{fig:gen_data}.
We visualized two-dimensional cost profiles of three samples of cost function parameters for 
each example.
The upper row shows cost functions of three high posterior probability samples when recovering the cost function from simulated behavioral response data, 
from left to right: the sample with the highest likelihood and, in order to portray the entire diversity and range of inferred cost functions, 
showing the sample with the highest and lowest inferred $\alpha$ value.
Low costs are shown in green while high costs increasingly fade into orange and white.\\

This two-dimensional visualization is especially useful since 
the one-dimensional visualization of the recovered shape of the cost functions in the second row of figure \ref{fig:gen_data} can be misleading and suggests a wider variety of potential cost functions despite the previously presented accurate parameter recovery. 
However this 
can be explained as an interplay of two factors: First, the high ground truth value of the $\alpha$ coefficient of $8$ causing a cost landscape which is very flat close to the target and has an extremely steep increase of costs for further deviations. 
Thus, differences in the $\beta$ parameter 
become more noticeable within close proximity to the target, 
but show only a small overall influence due to the cost function being dominated by alpha at larger distances.
With the second factor, the action variability $\sigma_a$ of $.3$, 
these small differences in costs at small deviations 
in A-C) of figure \ref{fig:costInf_2d_cost-tiles} 
become irrelevant.
Since especially the cost functions in A-C) have high $\alpha$ parameters,
cost values increase rapidly with larger deviations, 
leading to 
a difficult interpretability of short deviations 
relative to each other, see upper left and lower right 
of the two-dimensional cost tile plots in the middle row where 
deviations of actions from the target are maximal.
In order to solve this and make it visually more accessible
we also depicted the resulting tile plots with 
logarithmized values of cost in the last row.
There the high functional similarity of these cost functions becomes apparent again.
When in the end the action variability $\sigma_a$ of $.3$ is additionally considered
even the slight differences in costs for small deviations 
are no longer a relevant factor.\\

\subsection{Inference for continuous action control tasks}
In addition to generic data and theoretical statements, we will analyze anonymized data from two different continuous action-control tasks in the following. For this purpose, we infer the sensory and motor descriptive parameters and the cost functions on an individual level for two exemplary subjects each. In one task participants were asked to propel a puck pressing a key on a keyboard into gliding towards a target \parencite{neupaertl2020intuitive} and in the other they were asked to throw a bean bag with their hands to a target \parencite{willey2018long}. Both tasks are characterized by actions being on a continuum and both perception and the action itself being subject to variability. In the puck task however people could control the initial velocity of the object by the duration of a key press with uniformly drawn distances to the target to cover whereas in the beanbag task the object was directly controlled via muscle power for five different distances.\\

\begin{figure}[H]
	\centering
	\includegraphics[width=\textwidth]{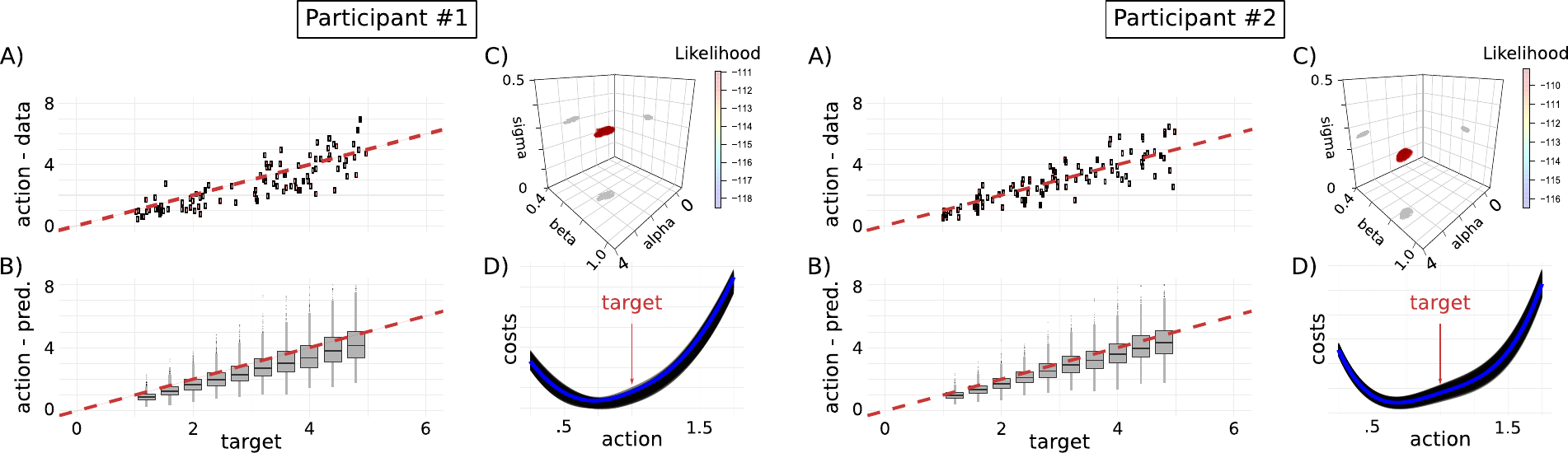}
	\caption{Puck sliding as action control task. Both targets and actions are on a continuous scale. (A) actual responses (y-axis) given the targets (x-axis). (B) predictions based on best inferred parameter setting as boxplots for exemplary targets (1,000 data points each). (C) most likely $5\%$ of posterior distribution for $\alpha$, $\beta$ and $\sigma_a$ in 3D parameter space. (D) cost functions corresponding to parameters in (C). Best sample highlighted in blue.}
	\label{fig:puck_data_inference}
\end{figure}

Inferred parameters for two participants of the puck task are shown in figure \ref{fig:puck_data_inference}: (A) showing the actual responses as a function of the target stimulus, (B) model predictions based on the most likely parameter combination of $\alpha$, $\beta$ and $\sigma_a$ (values for $\sigma_p$ were adopted from the original study), (C) the five percent sampled parameters with the highest likelihood in 3D space for better visibility and (D) the corresponding cost functions. Clearly, inferences for both subjects are similar. Participant one exhibits slightly higher variability in her responses (A) as inferred in (C). In both cases the convex shape of the cost function increasingly penalizes larger deviations.\\

\begin{figure}[!h]
	\centering
	\includegraphics[width=\textwidth]{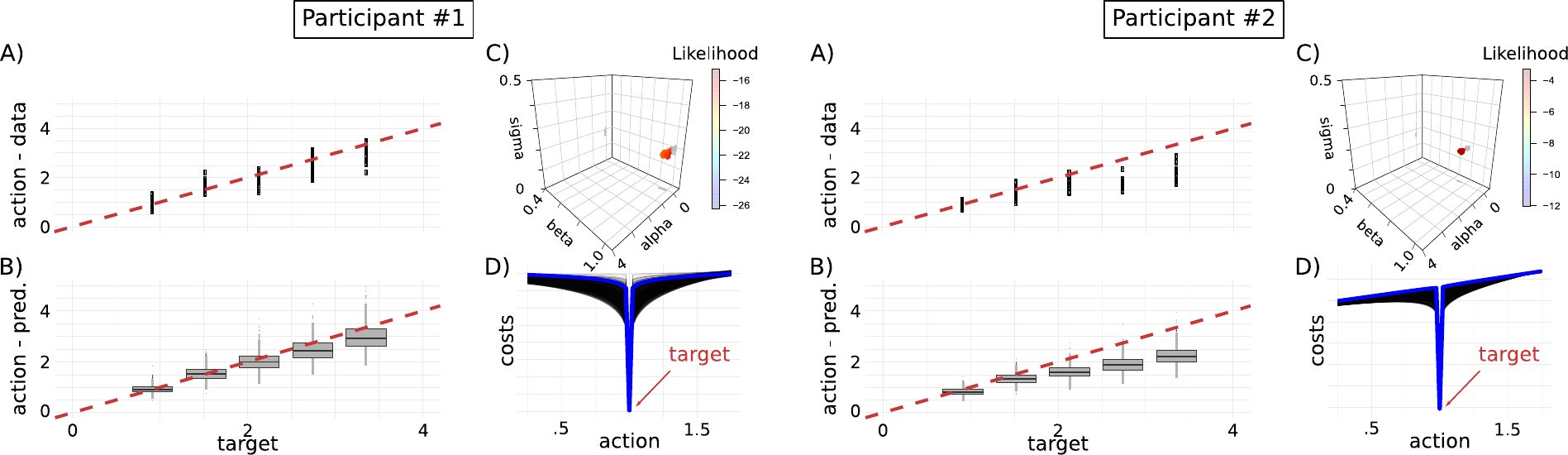}
	\caption{Beanbag throwing as action control task. Targets on five discrete positions and actions on a continuous scale. (A) actual responses (y-axis) given the targets (x-axis). (B) predictions based on best inferred parameter setting as boxplots for these targets (1,000 data points each). (C) most likely $5\%$ of posterior distribution for $\alpha$, $\beta$ and $\sigma_a$ in 3D parameter space. (D) cost functions corresponding to parameters in (C). Best sample highlighted in blue.}
	\label{fig:beanbag_data_inference}
\end{figure}

Figure \ref{fig:beanbag_data_inference} depicts responses and inferences for two participants of the beanbag throwing task: (A) showing the actual responses for each of the five target distances, (B) model predictions based on the most likely parameter combination of $\alpha$, $\beta$ and $\sigma_a$ ($\sigma_p$ again adopted from the original study), (C) the five percent sampled parameters with the highest likelihood in 3D space and (D) the corresponding cost functions. Again, we find strong similarities between subjects with respect to their action variability and shape of their cost function. Thereby participants in different tasks are potentially forming clusters in the 3D parameter space, and yet showing differences on the individual level. So that slight deviations in a parameter can lead to clear distinction in behavior, e.g. in figure \ref{fig:beanbag_data_inference} the lower value for $\beta$ in participant 2 causes a more asymmetric cost function and a stronger tendency to undershoot target values with increasing distance. However, here we showed only two examples as representatives for all participants from two action control tasks. Even though with them being the best individual representatives for the population they still do not reflect the whole diversity of parameters and thus cost functions and behavioral patterns.\\

\section{Discussion}
Production and reproduction tasks are a popular tool in psychophysics and yet also a complex computational problem, since people's behavior is influenced by diverse factors, for not only the perception but also the response in these continuous tasks are subject to variability while their behavior is additionally shaped by internally and externally motivated cost functions. Here we present an inference algorithm as a useful tool to investigate human behavior in these continuous decision making situations by inferring posterior beliefs about meaningful parameters describing the complete process from perception to final response. Using this framework we can explain seemingly suboptimal behavior, quantify responses in terms of meaningful parameters. In our framework we use logarithmic representation of perception and actions, naturally accommodating  Weber-Fechner phenomena in human behavior \parencite{battaglia2011haptic,neupaertl2020intuitive}. Additional support for the presence of such representations comes from recent studies that have indicated e.g. that intuitive representation of numbers in humans \parencite{siegler2003development,dehaene2008log} or time in animals \parencite{roberts2006evidence,yi2009rats} may be of logarithmic nature.\\

Similar work on a priori identifiability of probabilistic models in estimation tasks was done by \citet{acerbi_framework_2014}. They developed a probabilistic framework to recover prior beliefs in estimation tasks and to investigate candidate experimental designs a priori by comparing and ranking their identifiability. 
Here, we focus on inference of parameters that can describe perception, action, and multiple individual cost functions without the influence of strong priors over distributions in experiments. This can be particularly useful when distributions are complex, the experiment is too short to learn them, or there is no feedback on the actions at all. In doing so, we infer not only task specific costs but also, unlike \citet{acerbi_framework_2014}, the cost of performing the action itself. This part of the costs can be especially important for continuous tasks like production and reproduction tasks with increasing magnitude of the target stimulus, which can exemplary be seen for participant two in figure \ref{fig:beanbag_data_inference} with her increasing tendency to undershoot the target values. \\ 

Important prerequisites for our framework are the ability of subjects to assess their own uncertainty and variability and to adjust their behavior optimally given their uncertainty and cost functions. The high degree of the former capability was shown in studies investigating human movement planning and interval timing \parencite{sims2008adaptation,hudson2008optimal,balci2011optimal} and likewise for animals from rats \parencite{foote2007metacognition} to macaque monkeys \parencite{hampton2001rhesus}. The fact that subjects can adapt to external cost functions and the influence of costs, inherent in the execution of the action itself, has been shown in e.g. visuo-motor experiments where subjects explicitly adapt to external cost functions when pointing on a screen while also optimally include their own uncertainty \parencite{trommershauser2008decision}. However, here we additionally include costs for efforts. For example, it has been shown that people are sensitive to these costs in 2AFC tasks when the necessary force is manipulated \parencite{hagura2017perceptual}. These effort costs also need to be considered in computational models in line with bounded rationality to accurately describe human behavior as e.g. done for the suppression of blinking in an event detection task \parencite{hoppe2018humans}.\\

Given its trial by trial nature our proposed algorithm can be susceptible to outlier and inconsistent behavior. These data points can strongly alter inferred parameters, changing conclusions about perceptual uncertainty, action variability and cost functions. However, this vulnerability can be as well used to detect even subtle changes in behavior in sequential data. Taken together we introduced an useful inference tool to quantify behavior in continuous decision tasks and to a priori investigate appropriate experimental settings.n previous studies addressing differences in physical reasoning between tasks \parencite{smith2018different} and quantifying different sources of uncertainty in physical reasoning \parencite{smith2013sources} by adding the mode of physical interaction as an additional factor. This may also reconcile some previous result on intuitive physics, which reported strong deviations from Newtonian physics, but utilized very abstract depictions of scenes and no possibility for interaction \parencite{caramazza1981naive,todd1982visual}.\\

The present results are additionally relevant for the question of how the brain may implement physical reasoning. 
Previous studies have found evidence for the representation of abstract physical factors in parietal and frontal regions, when physics students thought about verbally presented physics terms \parencite{mason2016neural}. 
Similarly, recent studies involving physical reasoning about objects' dynamics on the basis of short movies also identified frontal and parietal regions 
representing abstract physical quantities such as mass \parencite{schwettmann2019invariant}
and involved in judging physical interactions \parencite{fischer2016functional}. These results give credence to the notion of causal generative models of physical objects and their interactions  compared to model-free pattern recognition approaches, such as those based on deep neural networks.
Nevertheless, the involvement of overlapping parietal regions in the representation of physical quantities such as mass when planning visuomotor interactions \parencite{gallivan2014representation} and the additional involvement of motor related regions in such tasks \parencite{chouinard2005role} speak for a crucial role of embodied representations  \parencite{anderson2003embodied,wilson2002six,foglia2013embodied} in physical reasoning  
at the implementational level.

\section{Acknowledgments}
This research was supported by “The Adaptive Mind”, funded by the Excellence Program of the Hessian Ministry of Higher Education, Science, Research and Art.
\printbibliography

\end{document}